\documentclass[10pt,twocolumn,letterpaper]{article}

\usepackage{cvpr}
\usepackage{times}
\usepackage{epsfig}
\usepackage{graphicx}
\usepackage{amsmath}
\usepackage{amssymb}
\usepackage{caption}
\usepackage{xcolor}
\usepackage{contour}
\usepackage{subcaption}

\usepackage{url}

\cvprfinalcopy % *** Uncomment this line for the final submission
\def\cvprPaperID{26} % *** Enter the CVPR Paper ID here

\ifcvprfinal\fi
\begin{document}

%%%%%%%%% TITLE
\title{SkelNetOn 2019: Dataset and Challenge on Deep Learning for \\ Geometric Shape Understanding \\ \small\url{http://ubee.enseeiht.fr/skelneton/}}
\author{
\.Ilke Demir$^1$, Camilla Hahn$^2$, Kathryn Leonard$^3$, Geraldine Morin$^4$, Dana Rahbani$^5$, \\ Athina Panotopoulou$^6$, Amelie Fondevilla$^7$, Elena Balashova$^8$, Bastien Durix$^4$, Adam Kortylewski$^{5,9}$ \\ \  \\ 
$^1$DeepScale, $^2$Bergische Universit\"at Wuppertal, $^3$Occidental College, $^4$University of Toulouse, \\ $^5$University of Basel, $^6$Dartmouth College, $^7$Universit\'e Grenoble Alpes, $^8$Princeton University,\\ $^9$ Johns Hopkins University\\}

\twocolumn[{%
\renewcommand\twocolumn[1][]{#1}%
\maketitle
\begin{center}
    \centering
    \includegraphics[width=1\textwidth]{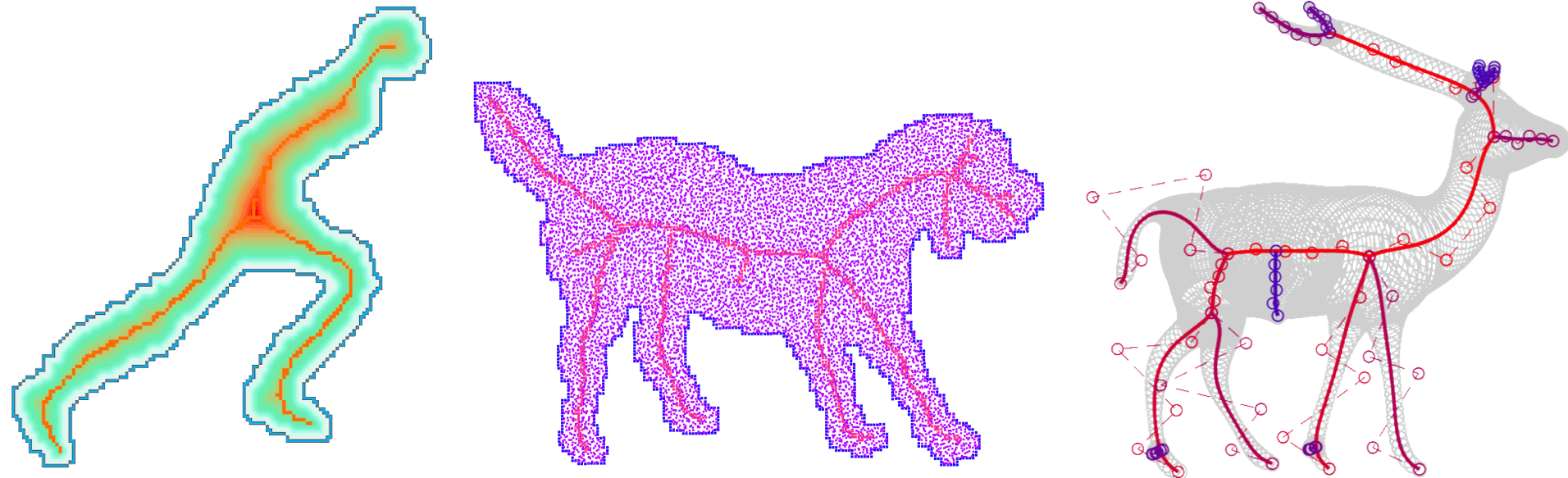}
    Figure 1: \textbf{SkelNetOn Challenges:} Example shapes and corresponding skeletons are demonstrated for the three challenge tracks in pixel (left), point (middle), and parametric domain (right).\label{fig:datas}
\end{center}%
}]
\setcounter{figure}{1}
\thispagestyle{empty}
%%%%%%%%% ABSTRACT
\begin{abstract}
We present \textit{SkelNetOn 2019 Challenge and Deep Learning for Geometric Shape Understanding} workshop to utilize existing and develop novel deep learning architectures for shape understanding. We observed that unlike traditional segmentation and detection tasks, geometry understanding is still a new area for deep learning techniques. SkelNetOn aims to bring together researchers from different domains to foster learning methods on global shape understanding tasks. We aim to improve and evaluate the state-of-the-art shape understanding approaches, and to serve as reference benchmarks for future research. Similar to other challenges in computer vision~\cite{davis18, coco}, SkelNetOn proposes three datasets and corresponding evaluation methodologies; all coherently bundled in three competitions with a dedicated workshop co-located with CVPR 2019 conference. In this paper, we describe and analyze characteristics of datasets, define the evaluation criteria of the public competitions, and provide baselines for each task.
\end{abstract}
\vspace{-25pt}

\section{Introduction}
% why competition
Public datasets and benchmarks play an important role in computer vision especially as the rise of deep learning approaches accelerate the relative scalability and reliability of solutions for different tasks. Starting with ImageNet~\cite{imagenet}, COCO~\cite{coco}, and DAVIS~\cite{davis18} for detection, captioning, and segmentation tasks respectively, competitions bring fairness and reproducibility to the evaluation process of computer vision approaches and improve their capabilities by introducing valuable datasets. Such datasets, and the corresponding challenges, also increase the visibility, availability, and feasibility of machine learning models, which brings up even more scalable, diverse, and accurate algorithms to be evaluated on public benchmarks.

% why shape understanding
Computer vision approaches have shown tremendous progress toward understanding shapes from various data formats, especially since entering the deep learning era. Although detection, recognition, and segmentation approaches achieve highly accurate results, there has been relatively less attention and research dedicated to extracting topological and geometric information from shapes. However, geometric skeleton-based representations of a shape provide a compact and intuitive representation of the shape for modeling, synthesis, compression, and analysis. Generating or extracting such representations is significantly different from segmentation and recognition tasks, as they condense both local and global information about the shape, and often combine topological and geometrical recognition.

% why skeletons
We observe that the main challenges for such shape abstraction tasks are (i) the inherent dimensionality reduction from the shape to the skeleton, (ii) the domain change as the true skeletal representation would be best expressed in a continuous domain, and (iii) the trade off between the noise and representative power for the skeleton to prohibit overbranching but still preserve the shape. Although the lower dimensional representation is a clear advantage for shape manipulation, it raises the challenge of characterizing features and representation, especially for deep learning. Computational methods for skeleton extraction are abundant, but are typically not robust to noise on the boundary of the shape (see \cite{Tagliasacchi2016}, for example). Small changes in the boundary result in large changes to the skeleton structure, with long branches describing insignificant bumps on the shape. Even for clean extraction methods such as Durix's robust skeleton \cite{Durix2019} used for our dataset, changing the resolution of an image changes the value of the threshold for keeping only the desirable branches. Training a neural network to learn to extract a clean skeleton directly, without relying on a threshold, would be a significant contribution to skeleton extraction. In addition, recent deep learning algorithms have shown great results in tasks requiring dimensionality reduction and such approaches could be easily applied to the shape abstraction task we describe in this paper.

Our observation is that deep learning approaches are useful for proposing generalizable and robust solutions since classical skeletonization do lack robustness. Our motivation arises from the fact that such deep learning approaches need comprehensive datasets, similar to 3D shape understanding benchmarks based on ShapeNet~\cite{shapenet}, SUNCG~\cite{suncg}, and SUMO~\cite{sumo} datasets, with corresponding challenges. The tasks and expected results from such networks should also be well-formulated in order to evaluate and compare them properly. We chose skeleton extraction as the main task, to be investigated in pixel, point, and parametric domains in increasing complexity. 

% introducing SkelNetOn
In order to solve the proposed problem with deep learning and direct more attention to geometric shape understanding tasks, we introduce SkelNetOn Challenge. We aim to bring together researchers from computer vision, computer graphics, and mathematics to advance the state of the art in topological and geometric shape analysis using deep learning. The datasets created and released for this competition will serve as reference benchmarks for future research in deep learning for shape understanding. Furthermore, different input and output data representations can become valuable testbeds for the design of robust computer vision and computational geometry algorithms, as well as understanding deep learning models built on representations in 3D and beyond. The three SkelNetOns are defined below:
\begin{itemize}
\item \textbf{Pixel SkelNetOn:}
As the most common data format for segmentation or classification models, our first domain poses the challenge of extracting the skeleton pixels from a given shape in an image. 
%The participants need to overcome fundamental problems like class imbalance, global structure search, and robustness constraints while reducing the given shapes to clean skeleton pixels. Although the output will not be a true geometric representation, it is easier to convert the skeleton pixels to a vector format. We expect the challengers to provide results in terms of the accuracy better than the current best skeleton extraction from images in the system. This will be a binary classification problem to detect the skeleton pixels for a given shape image. The evaluation will be based on completeness, smoothness, and connectedness of the skeleton pixels.

\item \textbf{Point SkelNetOn:} 
The second challenge track investigates the problem in the point domain, where the shapes will be represented by point clouds as well as the skeletons. 
%This track also emphasizes some fundamental questions as how to process non-uniform data, how to overcome class imbalance, and maybe some exploration in higher dimensional point clouds. We expect the challengers to provide results in terms of the accuracy better than the current best skeleton extraction from points in the system. This can be posed as a binary classification problem to assign a skeleton/non-skeleton class to all points in the given point cloud; however other formulations (i.e., as in transformer networks) are also welcome to solve this challenge. The evaluation will be based on the compactness and completeness of the skeleton points.

\item \textbf{Parametric SkelNetOn:} 
The last domain aims to push the boundaries to find parametric representation of the skeleton of a shape, given its image. The participants are expected to output skeletons of shapes represented as parametric curves with a radius function. 
 %The main challenge of this track arises from the domain change between the input and output, so representation of the output in a deterministic way is the key motivation of this track. We expect the challengers to provide results in terms of the accuracy better than the current best parametrized skeleton. This will be a recognition problem (similar to the problem of pose estimation) to detect the geometric representation (Bezier curves) for a given shape image. The evaluation will be based on the consistency, deviation, and accuracy of the parametrized skeleton representation evaluated by metrics on the induced shape boundary.
\end{itemize}

% info about sections and workshop ads 
In the next section, we introduce how the skeletal models are generated. We then inspect characteristics of our datasets and the annotation process (Section~\ref{sec:data}), give description of the tasks and formulations of evaluation metrics (Section~\ref{sec:tasks}), and introduce state-of-the-art methods as well as our baselines (Section~\ref{sec:res}). The results of the competition will be presented in the Deep Learning for Geometric Shape Understanding Workshop during the 2019 International Conference on Computer Vision and Pattern Recognition (CVPR) in Long Beach, CA on June 17th, 2019.  As of April 17th, more than 200 participants have registered in SkelNetOn competitions and there are 37 valid submissions in the leaderboards over the three tracks. Public leaderboards and the workshop papers are listed in our website\footnote{http://ubee.enseeiht.fr/skelneton/}.

\section{Skeletal Models and Generation}\label{sec:skeletons}
The Blum medial axis (BMA) is the original skeletal model, consisting of a collection of points equidistant from the boundary (the skeleton) and their corresponding distances (radii)~\cite{Blum1973}. The BMA produces skeleton points both inside and outside the shape boundary. Since each set, interior and exterior, reconstructs the boundary exactly, we select the interior skeleton to avoid redundancy. For the interior BMA, skeleton points are centers of circles maximally inscribed within the shape and the radii are their associated circles radii. For a discrete sampling of the boundary, Voronoi approaches to estimating the medial axis are well-known, and proven to converge to the true BMA as the boundary sampling becomes dense~\cite{ogniewicz1994}. In the following, we refer to the skeleton points and radius of the shape together as the interior BMA. 

The skeleton offers an intuitive and low dimensional representation of the shape that has been exploited for shape recognition, shape matching, and shape animation. However, this representation also suffers from poor robustness: small perturbations of the boundary may cause long branches to appear that model only a small boundary change. Such are uninformative about the shape. These perturbations also depend on the domain of the shape representation, since the noise on the boundary may be the product of coarse resolution, non-uniform sampling, and approximate parameterization. Many approaches have been proposed to remove these uninformative branches from an existing skeleton~\cite{Attali1997,Chazal2005,Giesen2009}, whereas some more recent methods offer a skeletonization algorithm that directly computes a clean skeleton~\cite{Durix2019,Leborgne2014}. 

We base our ground truth generation on this second approach. First, we apply one-pixel dilation and erosion operations on the original image to close some negligible holes and remove isolated pixels that might change the topology of the skeleton. We manually adjust the shapes if the closing operation changes the shape topology. Then, the skeleton is computed with 2-, 4-, and 6-pixel thresholds. In other words, the Hausdorff distance from the shape represented by the skeleton to the shape used for the skeletonization is at most 2, 4, or 6 pixels ~\cite{Durix2019}. We then manually select the skeleton that is visually the most coherent for the shape from among the three approximations to produce a skeleton which has the correct number of branches. Finally, we manually remove some isolated skeleton points or branches if spurious branches still remain. 

We compiled several shape repositories~\cite{Bronstein2007,Bronstein2008,leonardshape16,Sebastian2004}
%, including the MPEG7 dataset\footnote{http://www.dabi.temple.edu/$\sim$shape/MPEG7/dataset.html}), \question{The MPEG-7 database is a subset of the Sebastian2004 database, so we don't need to cite it twice.}  
for our dataset with $1,725$ shapes in 90 categories. We used the aforementioned skeleton extraction pipeline for obtaining the initial skeletons, and created shapes and skeletons in other domains using the shape boundary, skeleton points, and skeleton edges. %\question{I don't understand the preceding sentence. The pipeline takes images and extracts the skeleton points first, then identifies the boundary points associated to each skeleton point. --KL. Changed the explanation -ilke} 

%\TODO{}{issues with the traditional BMA -- noise, pruning, etc}

%\TODO{}{Initial automatic clean skeleton generation along with image morphological changes (Geraldine or Kathryn), then manual corrections (Camilla), thresholds, and manual pruning...}

\section{Datasets}\label{sec:data}
We converted the shapes and their corresponding ground truth skeletal models into three representation domains: pixels, points, and B\'ezier curves. This section will discuss these datasets derived from the initial skeletonization. 

\subsection{Shapes and Skeletons in Pixels}
\label{sec:skelimage}
The image dataset consists of $1,725$ black and white images given in portable network graphics format with size $256\times256$ pixels, split into $1,218$ training images, $241$ validation images, and $266$ test images. %\question{Are validation and test swapped? In the files I got from codalab, there are 242 validation images. KL}answer: I double checked they appear to be 1219 training and 242 validation images(by validation images I mean the test images inside the starting kit) I do not know about the test set since I do not know where to find it. 

\begin{figure}[hbt!]
	\centering
   		\includegraphics[width=1\linewidth]{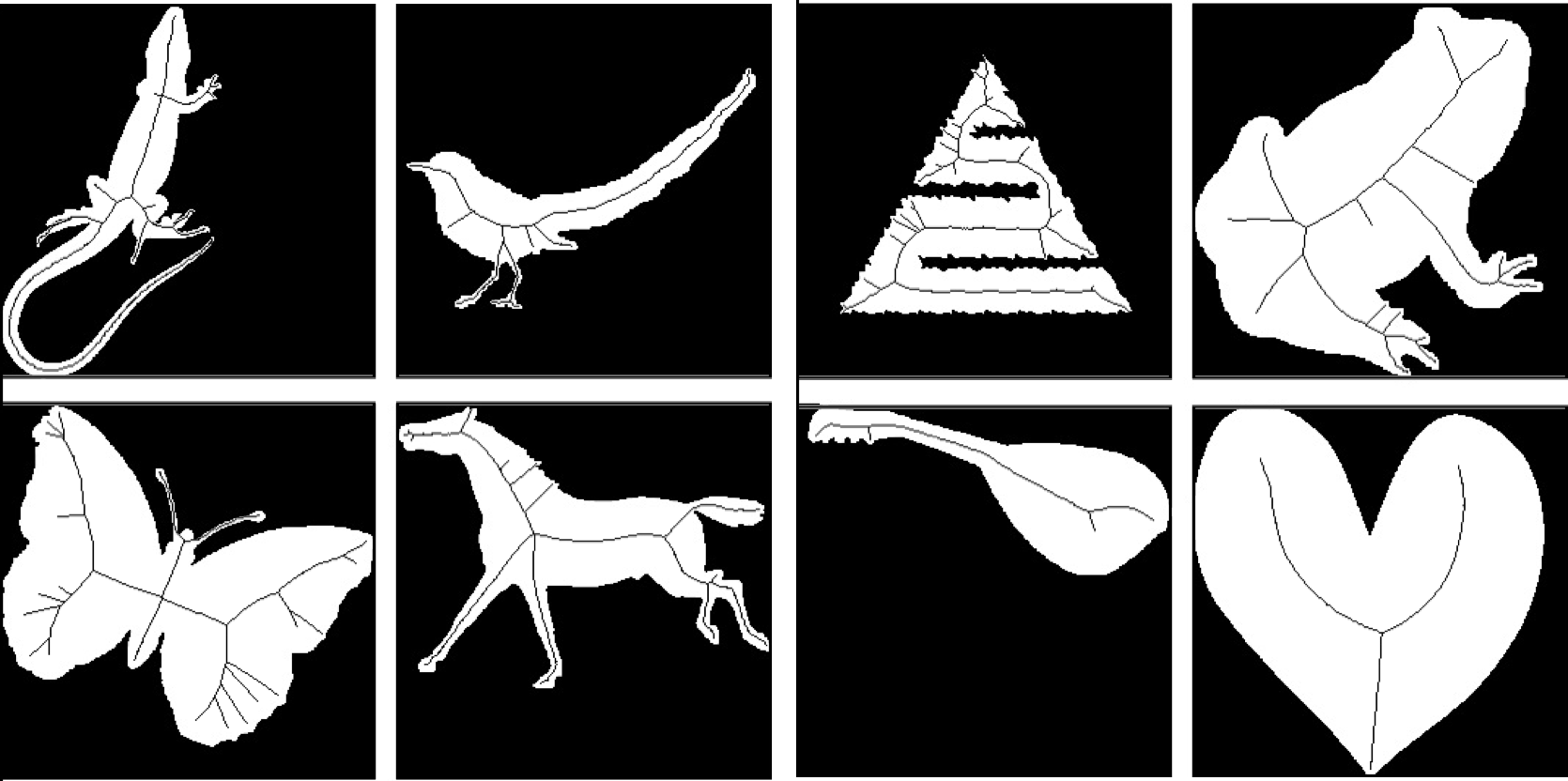}
   	\caption{\textbf{Pixel Dataset.} A subset of shape and skeleton image pairs is demonstrated from our pixel dataset. %We display negative skeleton images with enhanced edges for illustration purposes.
   	} \label{fig:pixels}
   \end{figure}

We provide samples from every class in both the test and validation sets. There are two types of images: the shape images which represent the shapes in our dataset (Figure~\ref{fig:pixels}), and the skeleton images which represent the skeletons corresponding to the shape images (Figure~\ref{fig:pixels}). In the shape images, the background is annotated with black pixels and the shape with a closed polygon filled with white pixels. In the skeleton images, the background is annotated with black pixels and the skeleton is annotated with white pixels. The shapes have no holes; some of them are natural, while others are man-made. If one overlaps a skeleton image with its corresponding shape image, the skeleton will lie in the ``middle" of the shape (i.e., it would be an approximation of the shape's skeleton). 
 
For generating shape images, inside of the shape boundaries mentioned in Section~\ref{sec:skeletons} are rendered in white, whereas outside is rendered in black. The renders are then cropped, padded, and downsampled to $256\times 256$. No noise was added or removed, therefore all expected noise is due to pixelation or resizing effects. For generating skeleton images, the skeleton points as well as all pixels linearly falling between two consecutive skeleton points are rendered in white, on a black background. By definition of the skeletons from the original generation process, we assume adjacency within 8-neighborhood in the pixel domain, and provide connected skeletons.
 
% \begin{figure}[hbt!]
%	\centering
%   	\begin{subfigure}{0.5\linewidth}
%   		\centering
%   		\includegraphics[width=0.95\linewidth]{img/sample_shapes.jpg}
%   		\caption{Shapes}
%   		\label{fig:dataset:shapes}
%   	\end{subfigure}%	
%   	\begin{subfigure}{0.5\linewidth}
%   		\centering
%   		\includegraphics[width=0.95\linewidth]{img/sample_skeletons_2.png}
%   		\caption{Skeletons}
%   		\label{fig:dataset:skeletons}
%   	\end{subfigure}
%   	\caption{\textbf{Pixel Dataset.} A subset of the shape and skeleton image pairs is demonstrated from our pixel dataset. We display the negative skeleton images with enhanced edges for illustration purposes.}
%   \end{figure}

\subsection{Shapes and Skeletons in Points}
\begin{figure}[hb]
   		\centering
   		\includegraphics[width=1\linewidth]{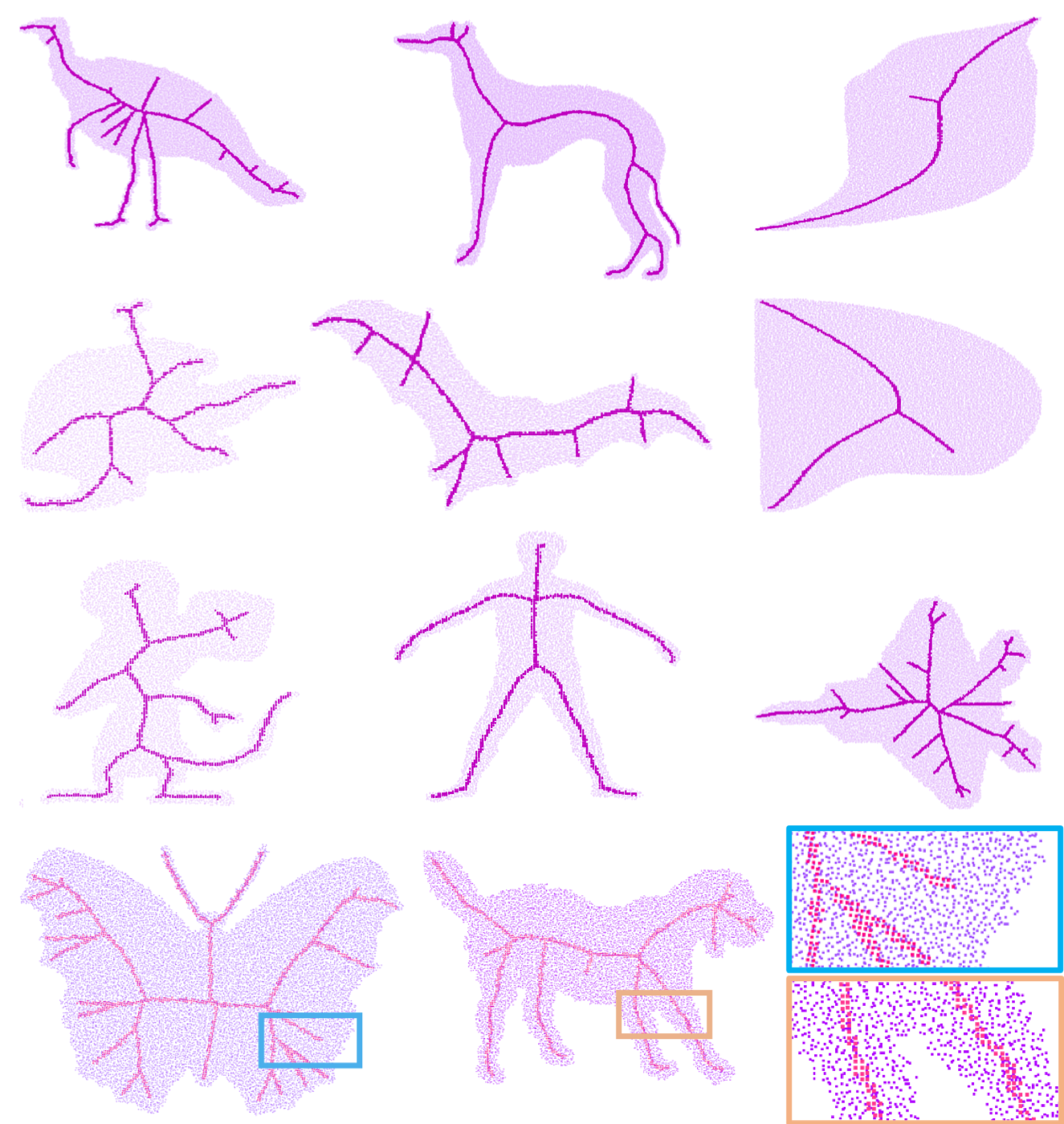}
   	\caption{\textbf{Point Dataset.} A subset of our shape and skeleton point cloud pairs is demonstrated. We also emphasize the point sampling using two close ups at the bottom right.}
   		\label{fig:dataset:points}
   \end{figure}

Similar to the image dataset, the point dataset consists of $1,725$ shape point clouds and corresponding ground truth skeleton point clouds, given in the basic point cloud export format \texttt{.pts}. Sample shape point clouds and their corresponding skeleton point clouds are shown in Figure~\ref{fig:dataset:points}.

The dataset is again split into $1,218$ training point clouds, $241$ validation point clouds, and $266$ test point clouds. We derive the point dataset by extracting the boundaries of the shapes (mentioned in Section~\ref{sec:skeletons}) as two-dimensional point clouds. We fill the closed region within this boundary by points that implicitly lie on a grid with granularity $h=1$. After experimenting with over/under sampling the point cloud with different values of $h$, we end up with this balancing value because the generated point clouds were representative enough to not lose details, and still computationally reasonable to process. Even though the average discretization parameter is given as $h=1$ in the provided dataset, we shared scripts\footnote{https://github.com/ilkedemir/SkelNetOn} in the competition starting kit to coarsen or populate the provided point clouds so participants are able to experiment with different granularities of the point cloud representation. To prevent the comfort of regularity which is observed in the pixel domain, we add some uniformly distributed noise and avoid any structural dependency in the later computed results. The noise is scaled by the sampling density, and we also provide scripts to apply noise with other probability distributions, such as Gaussian noise. 

Ground truth for the skeletal models are given as a second point cloud which only contains the points representing the skeleton. To compute the skeletal point clouds, we computed the proximity of each point in the shape point cloud to the skeleton points and skeleton edges from the original dataset. Shape points closer than a threshold (depending on $h$) to any original skeleton points or edges in Euclidean space are accepted for the skeleton point cloud. This generation process allows one-to-one matching of each point in the skeleton point cloud to a point in the shape point cloud, thus the ground truth can be converted to labels if the task in hand would be assumed as a segmentation task.

\subsection{Parametric Skeletons}\label{sec:skelprm}
Finally, the parametric skeleton dataset consists of $1,725$ shape images and corresponding ground truth parametric skeletons, exported in tab separated \texttt{.csv} format. The dataset is again split into $1,218$ training shapes, $241$ validation shapes, and $266$ test shapes. The shape images are created as discussed in Section~\ref{sec:skelimage}, and parametric skeletons are modeled as degree five B\'ezier curves. Each curve corresponds to a branch of the skeleton, where the first two coordinates describe the $\{x,y\}$ location in the image of an inscribed circle center, and the third coordinate is the radius $r$ associated with the point. Output is a vector containing 3D $(x, y, r)$ coordinates of the control points of each branch.
\begin{equation}
    v = [x_0^0, y_0^0, r_0^0, x_1^0, y_1^0, r_1^0, .. x_5^0, y_5^0, r_5^0, x_0^1, y_0^1, r_0^1 ..],
\end{equation}
where $b_i^j = (x_i^j,y_i^j,r_i^j)$ is the $i$-th control point of the $j$-th branch in the medial axis.

From the simply connected shapes of the dataset mentioned in Section~\ref{sec:skeletons}, we first extract a clean medial axis representation. For a simply connected shape, the skeleton is a tree, whose joints and endpoints are connected by curves, which we call proto-branches. Unfortunately, the structure of the tree is not stable. Because skeletons are unstable in the presence of noise, a new branch due to a small perturbation of the boundary could appear and break a proto-branch into two branches. Moreover the tree structure gives a dependency and partial order between the branches, not a total order. To obtain a canonical parametric representation of the skeleton, we first merge branches that have been broken by a less important branch. We then order the reduced set of branches according to branch importance. For both steps, we use a salience measure, the Weighted Extended Distance Function (WEDF) function on the skeleton~\cite{leonardshape16}, to determine the relative importance of branches. The WEDF function has been shown to measure relative importance of shape parts in a way that matches human perception \cite{carlier2016}. 

\begin{figure}[hbt!]
   		\centering
   		\includegraphics[width=0.5\linewidth]{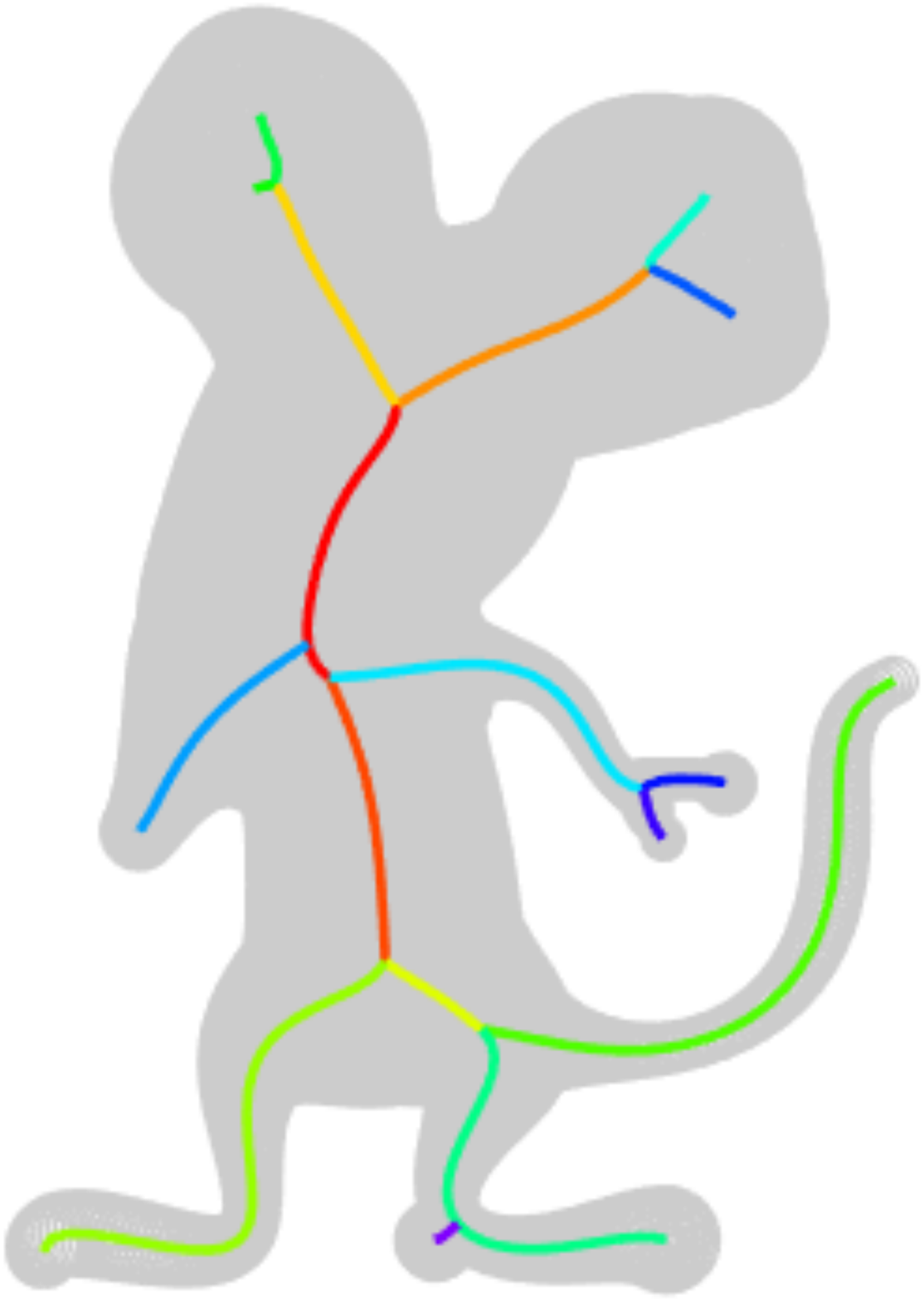}
   		\caption{\textbf{Skeletal Branches.} The red curve passes continuously through the arm and nose curves. However, legs (bottom) and ears (top) split the main red curve into two parts of equal importance, becoming the end-points. }
   		\label{fig:parametric}
\end{figure}

\paragraph{Merging branches.} 
First, we identify  pairs of adjacent branches that should be joined to represent a single curve: a branch is split into two parts at a junction induced by a child branch of lower importance if the WEDF function is continuous across  the junction. When two child branches joining the parent branch are of equal importance, then the junction is considered an end point of the parent curve. Figure~\ref{fig:parametric} shows the resulting curves.

\paragraph{Computation of the B\'ezier approximation.} 
Each individual curve resulting from the merging process is approximated by a B\'ezier curve of degree five, whose control points have three parameters $(x,y,r)$, where $(x,y)$ are coordinates, and $r$ is the radius. The end points of the curve are interpolated, and the remaining points are determined by a least square approximation. 

\paragraph{Branch ordering}
We then order the branches by importance to have a canonical representation. We estimate a branch importance by the maximal WEDF value of a point in the branch. The branches can then be ordered, and their successive list of control points is the desired output.

\section{Tasks and Evaluations}\label{sec:tasks}
Although there are different choices of evaluation metrics specific for each data modality, we formulate our metrics in accordance with the tasks for each domain.

\subsection{Pixel Skeleton Classification}
Generating skeleton images from the given shape images can be posed as a pixel-wise binary classification task, or an image generation task. This makes it possible to evaluate performance by comparing a generated skeleton image, pixel by pixel, to its ground truth skeleton image. Such a comparison automatically accounts for common errors seen in skeleton extraction algorithms such as lack of connectivity, double-pixel width, and branch complexity. 

However, using a simple $L1$ loss measurement would  provide a biased evaluation of image similarity. One can see this by looking at any of the skeleton images: the black background pixels far outnumber the white skeleton ones, giving the former much more significance in the final value of the computed $L1$ loss. To minimize the effects of class imbalance, the evaluation is performed using the $F1$ score, which takes into account the number of skeleton and background pixels in the ground truth and generated images. This is consistent with metrics used in the literature and will enable further comparisons with future work~\cite{shen2017deepskeleton}. The number of skeleton pixels (positive) or background pixels (negative) is first counted in both the generated and ground truth skeleton images. The $F1$ score is then calculated from the harmonic average of the precision and recall values as follows: 
 \begin{eqnarray}
 	F1 = \dfrac{2 \times precision \times recall } { precision + recall},
 \end{eqnarray}
using 
\begin{eqnarray}
    Precision = \dfrac{TP} {TP + FP} \nonumber \\
    Recall = \dfrac{TP}{TP + FN},
\end{eqnarray}
where $TP$, $FN$, and $FP$ stand for number of pixels for true positives, false negatives, and false positives respectively. 

\subsection{Point Skeleton Extraction}
Given a 2D point set representing a shape, the goal of the point skeleton extraction task is to output a set of point coordinates corresponding to the given shape's skeleton. This can be approached as a binary point classification task or a point generation task, both of which end up producing a skeleton point cloud that approximate the shape skeleton. The output set of skeletal points need not be part of the original input point set. The evaluation metric for this task needs to be invariant to the number and ordering of the points. The metric should also be flexible for different point sampling distributions representing the same skeleton. Therefore, the results are evaluated using the symmetric Chamfer distance function, defined by:
\begin{eqnarray}
Ch(A,B) = \frac{1}{|A|}\sum_{a\in A} \min _ {b\in B} ||a-b||_2 + \nonumber \\  \frac{1}{|B|}\sum_{b\in B} \min _ {a\in A} ||a-b||_2,
\end{eqnarray}
where $A$ and $B$ represent the skeleton point sets to be compared, $|.|$ denotes set cardinality, and $||.||$ denotes the Euclidean distance between two points. We use a fast vectorized \texttt{Numpy} implementation of this function in order to compute Chamfer distances quickly in our evaluation script.

\subsection{Parametric Skeleton Generation}
The parametric skeleton extraction task aims to recover the medial axis as a set of 3D parametric curves from an input image of a shape. Following the shape and parametric skeleton notations introduced in Section~\ref{sec:skelprm}, different metrics can be proposed to evaluate such a representation. In particular, we can measure either the distance of the output medial axis to the ground-truth skeleton, or its distance to the shape described in the input image. Although the second method looks better adapted to the task, it does not take into account several properties of the medial axis. It would be difficult, for example, to penalize disconnected branches or redundant parts of the medial axis.

We evaluate our results by distance to the ground truth medial axes in our database, since the proposed skeletal representation in the dataset already guarantees the properties introduced above and in Section~\ref{sec:skelprm}, and are ordered in a deterministic order. We use the mean squared distance between the control points on the original and predicted branches as:
\begin{equation}
    MSD(b,\tilde{b}) = \frac{1}{6}\sum_{i=0}^5 \left(  (x_i-\tilde{x}_i)^2 + (y_i-\tilde{y}_i)^2 + (r_i-\tilde{r}_i)^2 \right),
\end{equation}
where $b = {(x_i,y_i,r_i)}_{i=\{0..5\}}$ is a branch in the ground truth data, and $\tilde{b} = {(\tilde{x}_i,\tilde{y}_i,\tilde{r}_i)}_{i=\{0..5\}}$ is the corresponding branch in the output data.

The evaluation metric needs to take into account models with an incorrect number of branches, since this number is different for each shape. We penalize each missing (or extra) branch in the output with a measure on the length of the branch in the ground truth (or in the output). We use a measure called missing branch error (MBE) for each missing or extra branch $b$:
\begin{equation}
    MBE(b) = \frac{1}{5}\sum_{i=0}^4 (x_{i+1} - x_i)^2 + (y_{i+1} - y_i)^2
 + \frac{1}{6}\sum_{i=0}^5 r_i^2.
\end{equation}

Finally, the evaluation function between an output vector $\tilde{V}$ and its associated ground truth $V$ is defined as: 
\begin{equation}
D(V,\tilde{V}) = \frac{1}{N_b} \left( 
        \sum_{j=0}^{n_b-1} MSD(b^j,\tilde{b}^j) + 
        \sum_{j=n_b}^{N_b-1} MBE(\hat{b}^j)
        \right )
\end{equation}
where $N_b,n_b$ are respectively the number minimal and maximal of branches in the ground truth and in the output, and $\hat{b}$ are branches in the vector containing the maximal number of branches.

\section{State-of-the-art and Baselines}\label{sec:res}
Skeleton extraction has been an interesting topic of research for different domains. We briefly introduce these approaches in addition to our own deep learning based baselines. The participants are expected to surpass these baselines in order to be eligible for the prizes. 

\subsection{Pixel Skeleton Results}
%\paragraph{Related work.} 
Early work on skeleton extraction was based on using segmented images as input \cite{bai2007skeleton}. A comprehensive overview of previous approaches using pre-segmented inputs is presented in \cite{saha2016survey}. With the advancement of neural network technology, the most recent approaches perform skeletonization from natural images with fully convolutional neural networks \cite{shen2017deepskeleton,shen2016object}. In this challenge, we consider the former type of skeletonization task using pre-segmented input images.

\begin{figure}[ht]
	\centering
   	\begin{subfigure}{0.5\linewidth}
   		\centering
   		\includegraphics[width=0.7\linewidth]{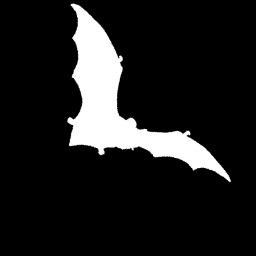}
   		\caption{}
   		\label{fig:dis-shape}
   	\end{subfigure}%	
   	\begin{subfigure}{0.49\linewidth}
   		\centering
   		\includegraphics[width=0.7\linewidth]{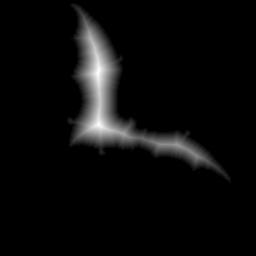}
   		\caption{}
   		\label{fig:dis-dis}
   	\end{subfigure}\\
   	\begin{subfigure}{0.5\linewidth}
   		\centering
   		\includegraphics[width=0.7\linewidth]{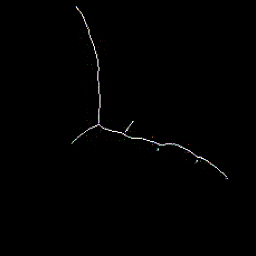}
   		\caption{}
   		\label{fig:dis-shape}
   	\end{subfigure}%	
   	\begin{subfigure}{0.49\linewidth}
   		\centering
   		\includegraphics[width=0.7\linewidth]{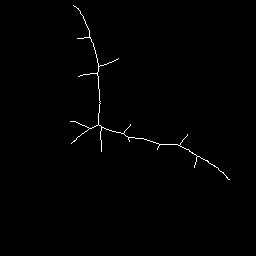}
   		\caption{}
   		\label{fig:dis-dis}
   	\end{subfigure}\\   	
	\caption{\textbf{Pixel SkelNetOn Baseline.} (a) Original shape image. (b) Distance transformed shape image. (c) Baseline prediction result. (d) Ground truth skeleton. }
	\label{fig:dis}
\end{figure}

%\paragraph{Our baseline result.} 
We formulate the task of skeletonization as image translation from pre-segmented images to skeletons with a conditional adversarial neural network. For our baseline result, we use a vanilla pix2pix %\footnote{https://github.com/phillipi/pix2pix}
model as proposed in~\cite{isola2017image}. We apply a distance transform to preprocess the binary input shape as illustrated in Figure~\ref{fig:dis}. We found that this preprocessing of the input data enhances the neural network learning significantly. The model is trained with stochastic gradient descent using the $L1$-image loss for $200$ epochs. We measure performance in terms of $F1$ score achieving a test accuracy of $\textbf{0.6244}$ on the proposed validation set.  

\subsection{Point Skeleton Results}
Skeleton and medial axis extraction from point clouds have been extensively explored in 2D and 3D domains by using several techniques such as Laplacian contraction~\cite{cao10}, mass transport~\cite{jalba16}, medial scaffold~\cite{leymarie07}, locally optimal projection~\cite{Lipman07}, maximal tangent ball~\cite{Ma2012}, and local $L_1$ medians~\cite{Huang13}. Although these approaches extract approximate skeletons, the dependency on sampling density, non-uniform and incomplete shapes, and the inherent noise in the point clouds are still open topics that deep learning approaches can handle implicitly. One such recent approach (P2PNet~\cite{p2pnet}) builds a Siamese architecture to build spatial transformer networks for learning how points can be moved from a shape surface to the medial axis and vice versa.

Based on the definition of this task in Section~\ref{sec:tasks}, it can be formulated as (i) a generation task to create a point cloud by learning the distribution from the original shapes and skeleton points, (ii) a segmentation task to label the points of a shape as skeletal and non-skeletal points, and (ii) a spatial task to learn the pointwise translations to transform the given shape to its skeleton. We chose the second approach to utilize state-of-the-art point networks. First, we obtained ground truth skeletons as labels for the original point cloud (skeletal label as $1$, and non-skeletal labeled as $2$). Then we experimented with PointNet++~\cite{pointnet} architecture's part-segmentation module to classify each pixel into two classes, defined by the labels. As the dimensionality of the classes are inherently different, we observed that the skeleton class collapses. Similarly, if we over-sample the skeleton, non-skeleton class collapses. To overcome this problem, we experimented with dynamic weighting of samples, so that the skeletal points are trusted more than non-skeletal points. The weights are adjusted at every $10^{th}$ epoch as well as the learning rate, to keep the balance of the classes. Our approach achieved an accuracy of $58.93\%$, measured by the mean IoU over all classes. Even though the mIoU is a representative metric for our task, we would like to evaluate our results better by calculating the Chamfer distance of skeletons in shape space in future.

\subsection{Parametric Medial Axis Results}

%\TODO{Can we have a brief related work on extracting B\'ezier skeletons here? (Geraldine)}{MISSING RELATED WORK}
Parametric representations of the medial axis have been proposed before, in particular representations with intuitive control parameters.  For example, Yushkevich. et al, \cite{yushkevich2003} use a cubic B-spline model with control points and radii; similarly Lam et al. \cite{lam2007} use piecewise B\'ezier curves.

We train a Resnet-50~\cite{He2016} modified for regression. Inputs are segmented binary images, and outputs are six control points in $\mathbb{R}^3$ that give $(x,y,r)$ values for the degree five B\'ezier curves used to model medial axis branches.

The parametric data is the most challenging of the three ground truths presented here, for two reasons. First, the number of entries in the ground truth varies with the number of branches in the skeleton. Second, the B\'ezier control points do not encode information about the connectivity of the skeleton graph. To overcome the first challenge, we set all shapes to have $5$ branches. For those shapes with fewer than $5$ branches, we select the medial point with the maximum WEDF value (a center of the shape) as the points in any non-existent branches, and set the corresponding radii to zero. We do not address the second issue for our baseline results. We use a simple mean squared error measure on the output coordinates as our loss function. While more branches are desirable to obtain quality models, there is a trade-off between the number of meaningless entries in the ground truth for a simple shape (with a low number of branches), and providing for an adequate number of branches in the ground truth of a complex shape (with many branches).
%For the second issue, we create a two-part custom loss function to penalize lack of graph connectivity. The first part computes MSE on a sampling of points from the branch curves (rather than on the control points themselves). The second part computes parameters of junction points on parent branches using the groundtruth, then takes the MSE over all junctions, penalizing the distance between the approximated branches at the associated groundtruth junction parameters and the starting points of the child branches. As the sum of convex functions, this loss function is also convex.

\begin{figure}
    \centering
    \includegraphics[width=\linewidth]{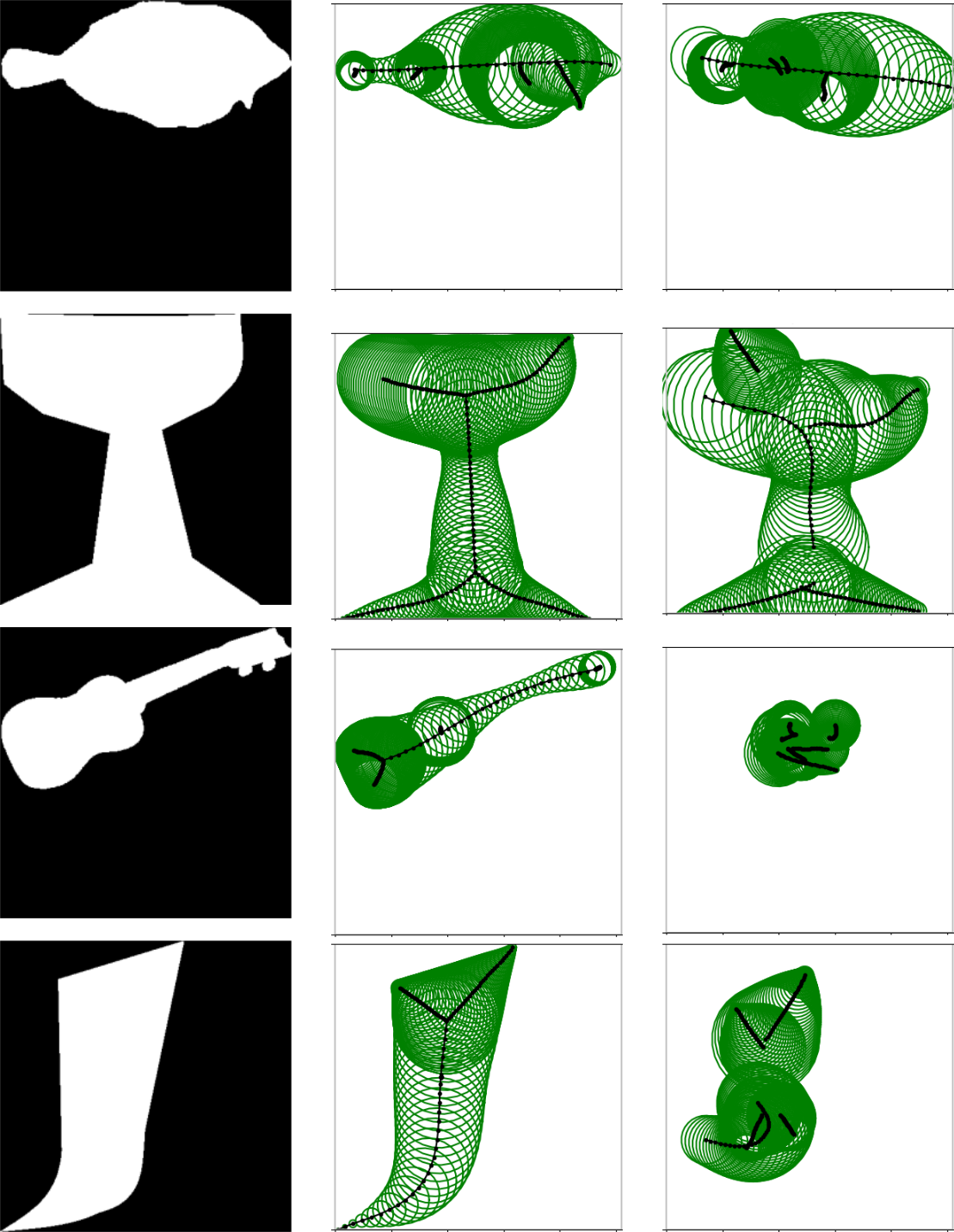}
    \caption{\textbf{Parametric SkelNetOn Baseline.} Input image (left), ground truth parametric medial axis (middle), prediction from our model (right).}
    \label{fig:paramres}
\end{figure}

We obtain a validation loss of $1405$ pixels, with a corresponding training loss of $121$ pixels, using early stopping. As a reference, the loss at the beginning of training is in the order of $50,000$ pixels.

As shown in Figure \ref{fig:paramres}, branch connectivity is only occasionally produced by our trained model. The model does well with location, orientation, and scale of shape, but often misses long and narrow appendages such as the guitar neck (third row). A future work is to incorporate other constraints in the architecture to encourage connectivity of branches. 

\section{Conclusions}
We presented SkelNetOn Challenge at Deep Learning for Geometric Shape Understanding, in conjunction with CVPR~2019. Our challenge provides datasets, competitions, and a workshop structure around skeleton extraction in three domains as Pixel SkelNetOn, Point SkelNetOn, and Parametric SkelNetOn. We introduced our dataset analysis, formulated our evaluation metrics following our competition tasks, and shared our preliminary results as baselines following the previous work in each domain. 

We believe that SkelNetOn has the potential to become a fundamental benchmark for the intersection of deep learning and geometry understanding. We foresee that the challenge and the workshop will enable more collaborative research of different disciplines on the crossroads of shapes. Ultimately, we envision that such deep learning approaches can be used to extract expressive parametric and hierarchical representations that can be utilized for generative models \cite{3dgan} and for proceduralization \cite{demir18}.
\section*{Acknowledgments}
Without the contributions of the rest of our technical and program committees, this workshop would not happen, so thank you everyone, in particular Veronika Schulze for her contributions throughout the project, Daniel Aliaga as the papers chair, and Bedrich Benes and Bernhard Egger for the support. The authors thank AWM's Women in Shape (WiSH) Research Network, funded by the NSF AWM Advances! grant. We also thank the University of Trier Algorithmic Optimization group, who provided funding for the workshop where this challenge was devised. We would also like to thank NVIDIA for being the prize sponsor of the competitions. Lastly, we would like to acknowledge the support and motivation of the CVPR chairs and crew. 
{\small
\bibliographystyle{ieee}
\bibliography{egbib}
}

\end{document}